\documentclass{Interspeech}

\interspeechcameraready

\title{Efficient Multilingual ASR Finetuning via LoRA Language Experts}

\author[affiliation={1,2}]{Jiahong}{Li$^*$}
\author[affiliation={2}]{Yiwen}{Shao}
\author[affiliation={2}]{Jianheng}{Zhuo}
\author[affiliation={1}]{Chenda}{Li}
\author[affiliation={2}]{Liliang}{Tang}
\author[affiliation={2}]{Dong}{Yu}
\author[affiliation={1}]{Yanmin}{Qian$^\dagger$}

\affiliation{}{}{$^1$Auditory Cognition and Computational Acoustics Lab}
\affiliation{}{}{MoE Key Lab of Artificial Intelligence, AI Institute}
\affiliation{}{}{School of Computer Science, Shanghai Jiao Tong University, China}
\affiliation{}{}{$^2$Tencent AI Lab, USA}

\email{\{li\_jiahong,yanminqian\}@sjtu.edu.cn}

\usepackage{comment}
\usepackage[table]{xcolor}
\usepackage{multirow}
\usepackage{makecell}

\begin{document}

\maketitle

\ifinterspeechfinal
    \let\thefootnote\relax\footnotetext{$^*$ This work was conducted during internship at Tencent.}
    \let\thefootnote\relax\footnotetext{$^\dagger$ Corresponding author.}
\fi

\begin{abstract}
    
    Recent advancements in deep learning have significantly enhanced multilingual automatic speech recognition (ASR) due to the development of advanced model architectures and available large-scale multilingual datasets. Despite that, multilingual ASR still suffers from the curse of multilinguality in that different languages tend to interfere with each other, making it difficult for the ASR model to identify multiple languages effectively while sharing model capacity across them. This paper proposes an efficient finetuning framework for customized multilingual ASR via prepared LoRA language experts based on Whisper. Through LoRA expert fusion or knowledge distillation, our approach achieves better recognition performance on target languages than standard fine-tuning methods. Experimental results demonstrate that the proposed models yield approximately 10\% and 15\% relative performance gains in language-aware and language-agnostic scenarios, respectively.

\end{abstract}
\section{Introduction}
As a crucial component of human-machine interaction, automatic speech recognition (ASR) focuses on transcribing speech signals into their corresponding written text. With the advent of deep learning techniques, ASR has achieved remarkable success in various real-world scenarios\cite{graves2006connectionist,graves2013speech,chan2016listen,kim2017joint,li2022recent}. Meanwhile, multilingual ASR, which necessitates a single system capable of recognizing multiple languages, has gained popularity due to the increasing demand for cross-lingual applications. Large-scale datasets\cite{DBLP:conf/interspeech/PratapXSSC20,conneau2023fleurs,ardila-etal-2020-common} have enabled the development of massive multilingual ASR foundation models, such as Whisper \cite{whisper}, Google USM \cite{usm}, and MMS \cite{mms}, which facilitate efficient finetuning of customized multilingual ASR models.

Multilingual ASR models encounter two primary challenges in effectively dealing with multiple languages. 
Firstly, language interference is inevitable when the languages are belong to different linguistic families or the training data is imbalanced in terms of quantity and quality. Previous works attempt to mitigate this issue by language-specific modules\cite{gaur2021mixture,zhou2022configurable,wanglanguage,li2024enhancing} or better sampling strategies\cite{kannan2019large,pineiromartin24_interspeech} for training data.
This challenge can be more complex when we consider the \textit{catastrophic forgetting} problem\cite{xutowards,kwok2024continual}, especially when language expansion\cite{khassanov2024extending,song24_interspeech} is involved.
Secondly, language identification is a hidden task inside multilingual ASR. While high performance is feasible with given language identity (LID), the performance can deteriorate dramatically when LID is absent\cite{liu2024parameter,yan2024improving}. This gap arises from the cascaded errors inherent in the implicit language identification and monolingual recognition pipeline.

Efficiency is another essential consideration in customized multilingual ASR. With the success of foundation models such as Whisper, fine-tuning has emerged as a popular method for developing downstream customized models. In this context, Low-Rank Adaptation (LoRA)\cite{lora} proves to be particularly effective regarding parameter scale and tuning time due to its low-rank modeling on the weight updates. Also, LoRA can be stored as separate residual modules, which avoids language interference when those languages are not finetuned. 
Knowledge distillation (KD)\cite{salinas2022knowledge,ferraz2024multilingual} is another way to accelerate the finetuning process with guidance from the teacher model, usually a model with similar architecture but larger capacity.

To this end, we introduce an efficient multilingual ASR finetuning framework based on LoRA language experts, facilitating fast adaptation from the base Whisper model to downstream customized multilingual ASR models. A LoRA language expert refers to the language-specific LoRA parameters trained for one single language, which effectively captures the monolingual knowledge from sufficient training data and can be prepared in advance. Through simple linear combination, shallow layers of LoRA experts can be merged as the shared encoder and language router, producing a new customized LoRA mixture of language experts (MoLE) model to handle language-agnostic multilingual speech.
Moreover, through layer-wise knowledge distillation, LoRA language experts can provide discriminative guidance for a multilingual LoRA student model, thereby efficiently enhancing the model performance.
In summary, the contributions of this paper are as follows:
\begin{enumerate}
    \item[-] We propose LoRA language experts to achieve improved language-aware recognition performance and apply the linear combination to merge different experts as LoRA MoLE for effective language-agnostic inference.
    \item[-] Utilizing layer-wise knowledge distillation, a multilingual LoRA student model can achieve superior performance in language-agnostic scenarios with a faster finetuning process.
\end{enumerate}
\section{Background}
\subsection{Whisper}
In this work, we adopt Whisper-medium\cite{whisper} as the multilingual foundation model for convenience, yet other foundation models should have similar behaviors.
Whisper\cite{whisper} is a typical encoder-decoder Transformer model designed for multiple speech processing tasks, including multilingual speech recognition, speech translation, language identification, and voice activity detection. Whisper takes speech input $\mathbf{X}$ in the form of an 80-dimensional log-mel spectrum, with a fixed duration of 30 seconds after trimming or padding. The encoder module transforms the input speech features into hidden representation $\mathbf{H}$, capturing essential information for speech recognition. 
The consecutive decoder module generates in an auto-regressive manner the transcribed text tokens $\mathbf{\hat{Y}}=[\hat{y}_1,\dots,\hat{y}_L]$ where $L$ denotes the total length of tokens. The current token is conditioned on the previous tokens together with the prompts $\mathbf{P}$. This process can be formulated as follows:
\begin{align}
    \mathbf{H} &= \text{AudioEncoder}(\mathbf{X}) \\
    \hat{y}_l  &= \text{TextDecoder}(\mathbf{P}, \hat{y}_{1:l-1}, \mathbf{H})
\end{align}

\subsection{LoRA}
LoRA\cite{lora} is an advanced technique originally designed to enhance the parameter-efficient finetuning of large pretrained models in the context of natural language processing (NLP) tasks. Instead of updating all the model parameters during finetuning, LoRA introduces low-rank matrices as the updated components for target weight matrices inside the model. In practice, for a weight matrix $\mathbf{W}_i\in \mathbb{R}^{d_1\times d_2}$, the new weight matrix $\mathbf{\hat{W}}_i$ after finetuning is modeled as \begin{align}
    \mathbf{\hat{W}}_i&= \mathbf{W}_i+\mathbf{\Delta} \mathbf{W}_i \\
    \mathbf{\Delta} \mathbf{W}_i &= \mathbf{B}_i \mathbf{A}_i
\end{align}
where $\mathbf{B}_i \in \mathbb{R}^{d_1\times r}$ and $\mathbf{A}_i \in \mathbb{R}^{r\times d_2}$ are trainable low-rank matrices after decomposition with rank $r \ll \text{min}(d_1,d_2)$. 
LoRA significantly reduces the training parameters when the backbone weights are frozen. 
After finetuning, LoRA parameters can be merged back to the original weights to reduces storage, or kept isolated to serve as independent residual modules for specific tasks.

\section{Methods}
In this paper, we propose an efficient finetuning framework that leverages monolingual LoRA language experts to facilitate the finetuning process of customized multilingual ASR models. 

\subsection{Task description}
Customized multilingual ASR models are designed to target a specific range of languages, encompassing both the base languages included during Whisper's pretraining phase and additional languages for language expansion.
Multilingual ASR can be decomposed into two primary tasks: language identification and monolingual speech recognition. While these two tasks are usually integrated to mitigate the risk of cascaded errors, it remains crucial to maintain balance between them.
The Whisper decoder treats different language IDs as special tokens and decodes them prior to the sequence of text tokens.

\subsection{LoRA language experts}
LoRA is an effective approach for multilingual ASR finetuning, not only due to its parameter-efficiency but also because LoRA parameters can be isolated from the backbone Whisper model. 
Such isolation allows LoRA to naturally preserve recognition performance across all base languages as separate residual modules, thereby avoiding the \textit{catastrophic forgetting} problem.
We refer to one set of monolingual LoRA parameters including $\mathbf{A}_{i,j}$ and $\mathbf{B}_{i,j}$ as a LoRA expert for the $j$-$th$ target language and an original Whisper weight matrix $\mathbf{W}_i$, which is finetuned on sufficient monolingual ASR data until convergence.
LoRA language experts perform better than a single multilingual model due to the curse of multilinguality, where different languages can interfere with each other, causing performance degradation.  
Information from various languages in the database can be effectively stored within the corresponding LoRA language experts, facilitating the tuning process of new customized multilingual ASR models.

\subsection{Language-agnostic LoRA MoLE}
Although LoRA language experts can achieve nearly optimal performance when LID is provided, they cannot perform language-agnostic inference.
To handle that, a language identification module is required to provide routing decisions among different experts.
Rather than training an independent classifier, we propose to combine shallow encoding layers of the prepared LoRA language experts to form shared multilingual layers. 
This leads to a LoRA mixture of language experts (MoLE) system, inspired from the MoE\cite{moe,loramoe} architecture. 

\begin{figure}
    \centering
    \includegraphics[width=\linewidth]{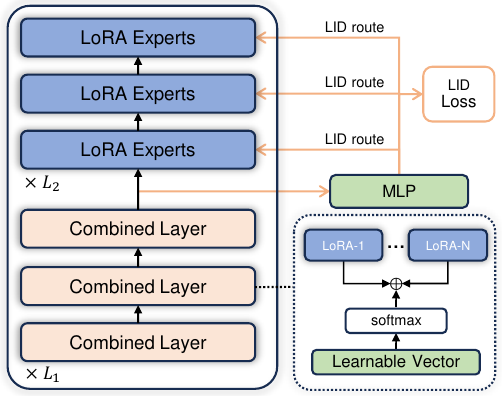}
    \caption{Paradigm of the MoLE encoder with combined LoRA language experts.}
    \label{fig:merge}
\end{figure}

As shown in Figure \ref{fig:merge}, we take $L_1$ encoding layers for linear combination, where all the weights of LoRA language experts are summed using learnable weights.
For $N$ LoRA experts associated with a specific weight matrix $\mathbf{W}_i$, a learnable vector $\mathbf{V}_i \in \mathbb{R}^{N}$ followed by a softmax layer is trained to generate the linear combination weights. The combination process can be formulated as follows:
\renewcommand{\arraystretch}{1.2}
\begin{align}
    \alpha_{i,j} &= \dfrac{\exp(\mathbf{V}_{i,j})}{\sum_k{\exp(\mathbf{V}_{i,k})}} \\
    \mathbf{\hat{A}}_i = \sum_{j}&{\alpha_{i,j}\mathbf{A}_{i,j}},\ \ 
    \mathbf{\hat{B}}_i = \sum_{j}{\alpha_{i,j}\mathbf{B}_{i,j}} \\
    \mathbf{\hat{W}}_i &= \mathbf{W}_i + \mathbf{\hat{B}}_i\mathbf{\hat{A}}_i
\end{align}
\renewcommand{\arraystretch}{1}

In this manner, multiple LoRA experts within those layers can be merged to reduce the total number of parameters and retain knowledge of all languages to produce LID routes with a simple multi-layer perceptron (MLP).
LoRA language experts and the Whisper backbone are kept frozen during finetuning of LoRA MoLE, with only the learnable vectors and the MLP updated based on the averaged LID loss and ASR loss.

\subsection{LoRA knowledge distillation}

\begin{figure}
    \centering
    \includegraphics[width=0.95\linewidth]{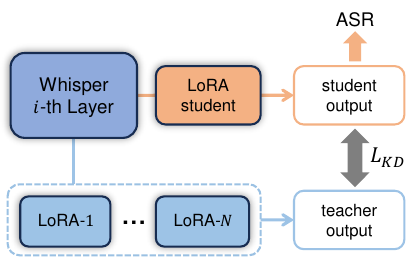}
    \caption{Process of layer-wise distillation from LoRA language experts to a multilingual LoRA student.}
    \label{fig:distill}
\end{figure}

Although LoRA MoLE is effective, it contained more model parameters due to the remaining expert modules as they cannot be merged to the backbone weight matrices. 
To obtain a simpler model, we propose to distill knowledge from the LoRA language experts into a single multilingual LoRA student.
Nonetheless, standard distillation on output ASR logits provides limited assistance on the fully-supervised finetuning process. 
Considering the residual format of LoRA, we assume the latent representations from different LoRA experts exhibit only minor differences. 
Consequently, a single multilingual LoRA student with a larger rank can align with multiple LoRA experts at the layer-wise granularity, and thus mimics the discriminative hidden distribution produced by the experts, which enhances the model's ability to distinguish languages.

As shown in Figure \ref{fig:distill}, we apply knowledge distillation to layer-wise output of the language-specific LoRA language experts and the language-agnostic LoRA student. 
To alleviate the cascaded mismatch in the forwarding path, the student output is interpolated with the teacher output before being passed to the next layer with a probability of 0.5. 
The layer-wise KD losses from different layers, including the output ASR logits, are averaged and combined with the ASR loss for backpropagation during training.
We use cosine similarity to compute the KD loss on the layer-wise outputs and Jensen divergence on the final ASR logits.
The process of knowledge distillation is follows: 
\begin{align}
    \mathbf{Y}_i &= \text{Layer}_i(\mathbf{Y}_{i-1})\\
    \mathbf{\hat{Y}}_i &= \text{Layer}_i(\mathbf{\tilde{Y}}_{i-1})\\
    \mathcal{L}_{\text{KD}_i}&=1-\frac{\mathbf{Y}_i\mathbf{\hat{Y}}_i}{||\mathbf{Y}_i|| \cdot ||\mathbf{\hat{Y}}_i||}\\
    \mathbf{\tilde{Y}}_i&=\frac12 (\mathbf{\hat{Y}}_i + \mathbf{Y}_i)\\
    \mathcal{L}&=\mathcal{L}_{\text{ASR}}+\alpha \frac{1}{M}\sum_{i=0}^{M}\mathcal{L}_{\text{KD}_i}
\end{align}
Here, $\alpha$ is the balancing weight, $M$ is the total number of layers, and $\mathbf{Y}_i,\mathbf{\hat{Y}}_i,\mathbf{\tilde{Y}}_i$ represent the teacher output, student output, and interpolated student output of the $i$-th layer, respectively.
During training, the Whisper backbone and LoRA language experts are kept frozen, and only the LoRA student will be updated, which is initialized from the average value of LoRA experts.

\section{Experiments}

\subsection{Setup}
The backbone model is Whisper-medium which is well-pretrained to produce reasonable transcriptions for a wide range of languages.
We select Zh (Chinese), En (English), Ko (Korean), Ja (Japanese), Ru (Russian), Vi (Vietnamese), and Id (Indonesian) as base languages and Yue (Cantonese) as language expansion. 
Experiments are conducted on an internal multilingual database with over 10,000 hours of data for Zh and En and around 5000 hours for other languages, which is sufficient for finetuning.
For testing, we adopt AISHELL\cite{aishell} and LibriSpeech\cite{librispeech} for Zh and En, and CommonVoice\cite{ardila-etal-2020-common} for the others.

We apply LoRA to the attention modules, including the querying, key and value matrices, and fully-connected layers across all layers in both the encoder and decoder, with a rank of 64 for monolingual LoRA experts and 256 for multilingual LoRA baseline and teacher model.
During training, we use AdamW\cite{adamw} optimizer with a peak learning rate of 1e-4 for LoRA models and 1e-5 for fully-finetuned models, and the dynamic batch size is set to 160 seconds.
The prepared monolingual LoRA language experts are trained to converge and stored for later use. 
The maximum training steps are set to 100,000 for other multilingual models and 20,000 for the LoRA MoLE model. 
All models are trained with 8 NVIDIA V100 32GB GPUs. 
During training, we remove Whisper's restriction of 30-second input to make it more efficient.
We apply a uniform sampling strategy for multilingual training to balance different languages.
During inference, we use greedy search to decode test data and apply no external language model. 
The performance is evaluated using the word error rate (WER) for each language.

\subsection{Results and analysis}

\begin{table*}[!ht]
\caption{Performance Comparison of Different Multilingual ASR Systems on Eight Languages in WER(\%).}
\label{tab:main}
\setlength{\tabcolsep}{1.3pt}
\renewcommand{\arraystretch}{1.1} 
\begin{tabular}{l|c|ccccccccc|ccccccccc}
\toprule\toprule
  \multicolumn{1}{c|}{\multirow{2}{*}{\textbf{Model}}} & \multicolumn{1}{c|}{\multirow{2}{*}{\textbf{\makecell[c]{\#Train\\Param}}}} & \multicolumn{8}{c}{\textbf{Language-Aware Test}} & \textbf{} & \multicolumn{8}{c}{\textbf{Language-Agnostic Test}} & \textbf{} \\
 ~  && \textbf{Zh} & \textbf{En} & \textbf{Ko} & \textbf{Ja} & \textbf{Id} & \textbf{Ru} & \textbf{Vi} & \textbf{Yue} & \textbf{avg} & \textbf{Zh} & \textbf{En} & \textbf{Ko} & \textbf{Ja} & \textbf{Id} & \textbf{Ru} & \textbf{Vi} & \textbf{Yue} & \textbf{avg} \\
\midrule
 \text{Vanilla}  & 0 & 9.20 & 3.65 & 5.12 & 15.75 & 12.89 & 9.78 & 18.95 & 38.90 & 14.28 & 9.20 & 3.65 & 5.24 & 16.95 & 17.91 & 10.60 & 23.12 & 50.32 & 17.12 \\
 \text{Fully FT} & \multirow{2}{*}{727M} & 5.24 & 4.41 & 5.34 & 17.21 & 9.88 & 9.74 & 16.68 & 9.68 & 9.77 & 5.24 & 5.02 & 5.69 & 16.83 & 12.20 & 9.93 & 20.90 & 9.68 & 10.69 \\
 \text{Curated Fully FT} & ~ & 4.34 & 3.03 & 5.55 & 15.86 & 9.87 & 9.16 & 13.87 & 7.74 & 8.68 & 4.34 & 5.02 & 6.19 & 16.46 & 13.19 & 9.82 & 19.20 & \underline{7.74} & 10.25 \\
\midrule
 \text{Mono. LoRA LE} &66M$*$8& \underline{3.59} & \underline{2.44} & \underline{4.54} & \underline{14.31} & \underline{8.93} & 8.79 & \underline{11.75} & \underline{7.48} & \textbf{7.73} & \multicolumn{9}{c}{-} \\
 \text{LoRA MoLE}  &1M& 4.07 & 2.54 & 4.65 & 14.39 & 10.51 & \underline{8.77} & 14.79 & 8.88 & 8.58 & \underline{4.07} & \underline{2.54} & \underline{5.07} & \underline{14.92} & 12.88 & \underline{8.93} & 17.25 & 8.88 & \textbf{9.32} \\
\midrule
 \text{Multi. LoRA}  & ~ & 5.34 & 2.93 & 5.37 & 15.58 & 10.28 & 10.25 & 18.37 & 9.64 & 9.72 & 5.34 & 4.88 & 5.73 & 16.22 & 13.10 & 10.59 & 26.54 & 9.68 & 11.51 \\
 \text{Multi. LoRA KD} & \multirow{-2}{*}{264M} & 4.85 & 2.84 & 4.77 & 15.31 & 9.74 & 9.50 & 14.63 & 8.25 & \textbf{8.74} & 4.85 & 4.67 & 4.78 & 16.29 & \underline{11.68} & 9.80 & \underline{17.22} & 8.33 & \textbf{9.70} \\\bottomrule
\end{tabular}
\end{table*}

\begin{figure}
 \centering
 \includegraphics[width=\linewidth]{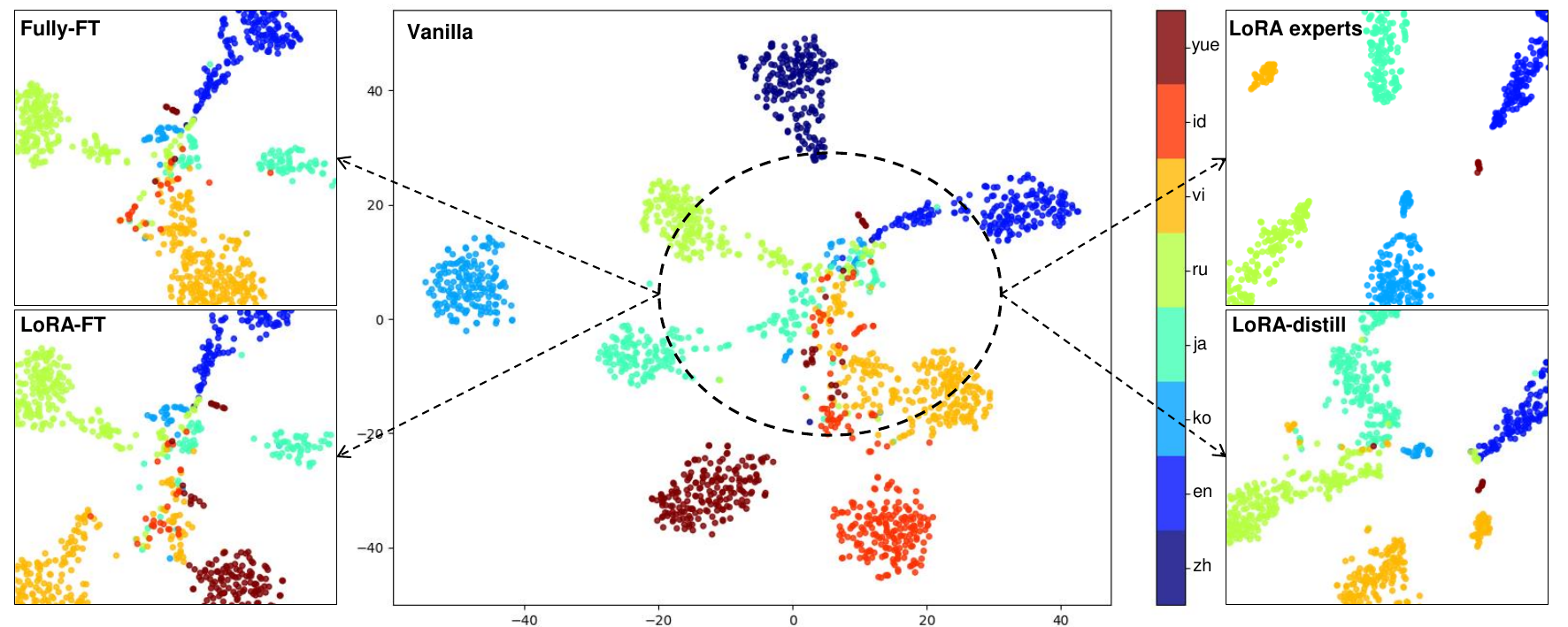}
 \caption{Visualization of the encoded hidden representations by t-SNE for different models.}
 \label{fig:dist}
\end{figure}

In Table \ref{tab:main}, we present the ASR performance of different multilingual ASR systems across eight languages. 
The vanilla Whisper model performs well on base languages but poorly on expanded Cantonese speech (using Zh for LID).
After fully-finetuning, the \textit{Fully-FT} model shows improved performance on Zh, Id, Vi and Yue, but performs worse on other languages, indicating interference between different languages.
The \textit{Curated Fully-FT} model is an enhanced version that uses four times the training steps and a curated language sampling strategy with weights of $\{1.0,1.0,0.25,0.25,0.25,0.25,0.25,0.5\}$ for those languages, as concluded from experiments. 
This model significantly improves performance on language-aware tests, particularly for En, Vi, and Yue, although the performance gain is not consistent in language-agnostic scenarios.

The \textit{Monolingual LoRA Experts} achieves the best performance with LID input due to their language independence, while it cannot perform language-agnostic inference with lack of language identification ability.
The proposed \textit{LoRA MoLE} model with 20 layers merged can remedy that drawback with the sacrifice of the monolingual performance and minimum training parameters. The average WER rises to 8.68\%, but this model provides the best average performance when LID input is unavailable.
Notably, the performance of MoLE is closely tied to language classification accuracy, which is adversely affected by language imbalance. Consequently, performance on relatively high-resource languages, such as Zh and En, aligns well with language-aware inference, while performance deteriorates for other languages, particularly Vi.

In the last block, the finetuned \textit{Multilingual LoRA} model performs worse than the fully-finetuned models, except for En, indicating a weaker modeling ability.
The proposed \textit{Multilingual LoRA KD} model exhibits approximately 10\% and 15\% relative performance gains on language-aware and language-agnostic tests, respectively,  due to the layer-wise alignment with LoRA language experts, demonstrating effective knowledge transfer.

In Figure \ref{fig:dist}, we visualize the hidden representations of different languages from five models using t-SNE\cite{tsne}.
It is evident that both the fully fine-tuned model and the LoRA model struggle to separate multilingual samples near the decision boundary, whereas the distilled LoRA student aligns well with the LoRA language experts, producing distinctly separable results.

\subsection{Ablation Study}

\begin{figure}
 \centering
 \includegraphics[width=\linewidth]{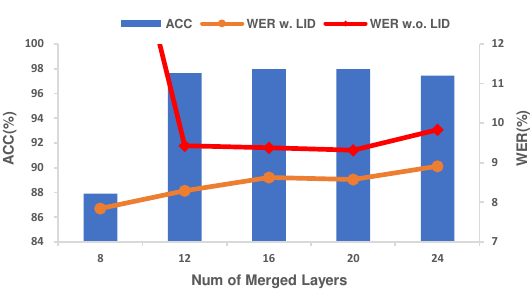}
 \caption{Variation of language routing accuracy and recognition WER regarding the number of merged layers in LoRA MoLE. {Whisper} encoder has 24 layers in total.}
 \label{fig:layers}
\end{figure}

\noindent\textbf{LoRA MoLE}\quad
In LoRA MoLE, the number of merged layers $L_1$ is a trade-off between the accuracy of language identification and the performance of remaining LoRA expert layers.
We evaluate the performance variation with $L_1\in\{8,12,16,20,24\}$, and present the results in Figure \ref{fig:layers}.
When $L_1=8$, the combination is too shallow to generate useful features for language identification. 
As more expert layers are merged, the ASR performance of the remaining layers declines when LID is known, while the identification accuracy peaks with 16 or 20 merged layers.
The best overall language-agnostic performance is achieved at $L_1=20$, which also significantly reduces the number of expert parameters within the encoder.

\begin{table}[ht]
\centering
\caption{Ablation Study on LoRA Knowledge Distillation in Averaged WER(\%) among Eight Languages.}
\label{tab:distill}
\setlength{\tabcolsep}{2pt}
\renewcommand{\arraystretch}{1.0} 
\begin{tabular}{l|c|c}
\toprule
\multicolumn{1}{l|}{\textbf{Models}} & \textbf{Language-Aware} & \textbf{Language-Agnostic} \\
\midrule
Multi. LoRA                             & 9.72                & 11.51                   \\
LoRA-KD                             & \textbf{8.74}                & \textbf{9.7}                   \\
LoRA-KD-rank128                     & 9.02                & 10.13                 \\
LoRA-KD-logits-only                 & 9.39                & 10.32                \\
\bottomrule
\end{tabular}
\end{table}

\vspace{5pt}
\noindent\textbf{LoRA KD}\quad
For LoRA KD, we compare the average WER with or without LID against other configurations, as shown in Table \ref{tab:distill}.
As we reduce the student rank from 256 to 128, the WER increases correspondingly but remains better than the basic multilingual LoRA.
Additionally, we apply the standard output-level KD on the ASR logits, which resulted in only minor performance enhancements.

\subsection{Limitation}
The total number of parameters for LoRA language experts increases as the number of target languages grows. 
This can lead to increased memory costs when training models with more languages.

\section{Conclusion}
In this paper, we introduce an efficient finetuning framework for customized multilingual ASR based on foundation models such as Whisper. 
By leveraging language-specific LoRA language experts, we can train a language-agnostic LoRA MoLE model at minimal cost, 
and achieve superior performance on both testing scenarios with a multilingual LoRA student through layer-wise knowledge distillation.

\ifinterspeechfinal
    \section{Acknowledgement}
This work was supported in part by China NSFC projects under Grants 62122050 and 62071288, in part by Shanghai Municipal Science and Technology Commission Project under Grant 2021SHZDZX0102, and in part by Tencent AI Lab Rhino-Bird Focused Research Program.

\fi

\bibliographystyle{IEEEtran}
\bibliography{mybib}

\end{document}